\pdfoutput=1

\documentclass[11pt]{article}

\usepackage[final]{acl}

\usepackage{times}
\usepackage{latexsym}

\usepackage[T1]{fontenc}

\usepackage[utf8]{inputenc}

\usepackage{microtype}

\usepackage{inconsolata}

\usepackage{graphicx}

\usepackage{enumitem}
\setlist[enumerate]{noitemsep, nolistsep}
\usepackage{amsmath}
\usepackage{amssymb}
\usepackage{cleveref}
\usepackage{booktabs}
\usepackage{siunitx}
\usepackage{multirow}
\usepackage{caption}
\usepackage{subcaption}

\title{To Aggregate or Not to Aggregate. That is the Question: A Case Study on
  Annotation Subjectivity in Span Prediction}

\author{
 \textbf{Kemal Kurniawan\textsuperscript{1}}\hspace{1em}
 \textbf{Meladel Mistica\textsuperscript{1}}\hspace{1em}
 \textbf{Timothy Baldwin\textsuperscript{1,2}}\hspace{1em}
 \textbf{Jey Han Lau\textsuperscript{1}}
\\
 \textsuperscript{1}The University of Melbourne\hspace{1em}
 \textsuperscript{2}MBZUAI
 \\
 \texttt{\{kurniawan.k,misticam,laujh\}@unimelb.edu.au}
 \\
 \texttt{tb@ldwin.net}
}

\newcommand{\random}{\textsc{Random}}
\newcommand{\mv}{\textsc{MV}}
\newcommand{\rel}{\textsc{ReL}}
\newcommand{\anon}[1]{#1}
\robustify\bfseries

\begin{document}
\maketitle
\begin{abstract}
  This paper explores the task of automatic prediction of text spans
  in a legal problem description that support a legal area label. We
  use a corpus of problem descriptions written by laypeople in English
  that is annotated by practising lawyers. Inherent subjectivity
  exists in our task because legal area categorisation is a complex
  task, and lawyers often have different views on a problem,
  especially in the face of legally-imprecise descriptions of
  issues. Experiments show that training on majority-voted spans
  outperforms training on disaggregated ones.\footnote{Code is available at
    \url{https://github.com/kmkurn/wassa2024}.}
\end{abstract}

\section{Introduction}

Automatic categorisation of lay descriptions of problems into relevant legal
areas is of critical importance for providers of free legal
assistance~\citep{mistica2021}. In our case, we have access to a dataset where a legal problem description is
annotated by multiple lawyers who first perform document-level annotation by
choosing relevant legal areas,\footnote{There are 32 possible legal areas
  including \textsc{Neighbourhood disputes}, \textsc{Elder law}, and
  \textsc{Housing and residential tenancies}.} and then, for each legal area chosen,
the lawyers highlight text spans that support their decision. These spans not
only help justify the selected areas of law but also improve the interpretability of
their decision.

The categorisation of legal areas is a complex problem, and lawyers sometimes have different views on a problem, especially when the task is performed on legally-imprecise descriptions of the personal circumstances of an individual.
Therefore, subjectivity is inherent in our task. This subjectivity leads to
annotation disagreements, both at the document- and the span-level. While such
disagreements are often seen as noise that needs to be eliminated in data
annotation~\citep{plank2022}, here they are signal because they are produced by
subject-matter experts.

In this paper, we explore the task of automatic span prediction using our
expert-annotated dataset, as illustrated in \Cref{tbl:annotated-ex}. Given a
problem description~(which is a short document) and an area of law, the task
aims to predict text spans in the description that support the area of law label.
We describe the development of machine learning models for the task that are
trained on a corpus containing legal problem descriptions written by laypeople
in English. Across various evaluation scenarios, we find that aggregating
training span annotations outperforms keeping them disaggregated.

\begin{table}\small
  \centering
  \begin{tabular}{@{}p{3cm}p{4cm}@{}}
    \toprule
    Area of law & Annotated description \\
    \midrule
    \textsc{Elder law} & I was admitted in a Public Hospital. I want the right to go home, \colorbox{orange}{NOT aged care!} \\
    \addlinespace
    \textsc{Guardianship and administration} & I was admitted in \colorbox{orange}{a Public Hospital.} I want the right to go home, \colorbox{orange}{NOT aged care!} \\
    \bottomrule
  \end{tabular}
  \caption{Examples of a description annotated with spans for two different
    areas of law.}\label{tbl:annotated-ex}
\end{table}

\section{Problem Statement}\label{sec:prblm-stmt}

Given a text expressed as a sequence of $N$ words $\boldsymbol{x}=x_1x_2\ldots x_N$
and a label $l$, the goal is to predict a set of non-overlapping spans
$S=\lbrace(b_i,e_i)\rbrace_{i=1}^M$ where $1\leq b_i\leq e_i\leq N$ such that
the text segments $\lbrace x_{b_i}x_{b_i+1}\ldots x_{e_i}\rbrace_{i=1}^M$
explain the reason for assigning $l$ to $\boldsymbol{x}$. In other words, $b_i$
and $e_i$ respectively denote the beginning and the end indices of the $i$-th
span supporting the assignment of $l$ to $\boldsymbol{x}$. We cast the problem
as sequence tagging by modelling the probability of $S$ given $\boldsymbol{x}$
and $l$ as
\begin{equation}
  P(S\mid\boldsymbol{x},l)\propto\exp f(\boldsymbol{x},\boldsymbol{y},l)\label{eqn:prob-spans}
\end{equation}
where $\boldsymbol{y}=y_1y_2\ldots y_N$ is a
sequence of $N$ tags representing the spans in $S$, each $y_i$ corresponds to $x_i$,
and $f$ is a real-valued function that measures the relevance of $\boldsymbol{y}$ in supporting the assignment of $l$ to $\boldsymbol{x}$. To get $\boldsymbol{y}$ from $S$,
we use an encoding where $y_i$ takes one of 5 possibilities depending on the
position of $i$ in a span~\citep{sekine1998}:
\begin{enumerate}
\item singleton, if $\exists(b,e)\in S$ where $b=e=i$;
\item beginning, if $\exists(b,e)\in S$ where $b=i<e$;
\item end, if $\exists(b,e)\in S$ where $b<i=e$;
\item inside, if $\exists(b,e)\in S$ where $b<i<e$; and
\item outside, otherwise.
\end{enumerate}
The span prediction problem is then equivalent to finding the highest scoring
sequence
\begin{equation*}
  \boldsymbol{y}^\star=\arg\max_{\boldsymbol{y}}f(\boldsymbol{x},\boldsymbol{y},l).
\end{equation*}
The sequence $\boldsymbol{y}^\star$ is then decoded to get the final predicted
spans.

\section{Corpus}

The corpus was collected by \anon{Justice
  Connect},\footnote{\url{https://\anon{justiceconnect.org.au}}} an
\anon{Australian public benevolent} institution\footnote{As defined by
  the \anon{Australian
    government:~\url{https://www.acnc.gov.au/charity/charities/4a24f21a-38af-e811-a95e-000d3ad24c60/profile}}}
that connects laypeople seeking legal assistance with pro bono
lawyers. On its website, \anon{Justice Connect} allows help-seekers to
describe their problem in free text format in English. After
anonymising identifiable information, problem descriptions collected
from July 2020 to early December 2023 were presented to a pool of
lawyers to be annotated. Each annotator selected one or more out of
the 32 areas of law that applied to the problem (thus it is a
\textit{multi-label} classification problem), representing the
different law specialisations the case relates to. On average, a problem
description is labelled with 3 areas of law. For each document-level area
of law selected, the annotator then select spans of words\footnote{The number of
  words must be at least three.} that support their decision.
On average, each problem description is annotated by 5
lawyers. This whole annotation process was carried out by \anon{Justice
  Connect}. In other words, we do not perform any additional annotation and
simply use the annotated corpus.

Relating to the problem statement in \Cref{sec:prblm-stmt}, the description and
the area of law form the inputs $\boldsymbol{x}$ and $l$ respectively, while the
spans make up the output $S$. Together, the input and the output form a labelled
example of the task. Following prior work on a similar
corpus~\citep{mistica2021}, we employ 20-fold cross validation to create the
training and the test sets and randomly take 10\% of the training set to form
the development set. Over the
20 folds we have a total of 35K unique problem description and legal area pairs, with a total of 3.8M words in the
problem descriptions.

\section{Method}

\subsection{Subjectivity-Aware Evaluation}\label{sec:subj-eval}

Because of the inherent subjectivity of the labelling task, a test
input~(consisting of a problem description and an area of law) can
have multiple valid span annotations whose boundaries may not match exactly.
Specifically for a given problem description, the same area of law can be
supported by different spans. Similarly, the same span can support different
areas of law. To deal with this mismatched boundaries issue, we adopt both span-
and word-level evaluation. To address the issue of multiple valid spans, we
experiment with 2 types of gold spans: majority-voted and best-matched. With
these strategies, we have a total of 4 combinations of evaluation setup.

\subsubsection{Span- and Word-Level Evaluation}

In span-level evaluation, a predicted span is considered correct if it starts
from and ends at the same positions as a gold span. In other words, their span
boundaries must match exactly to be considered equal.

In contrast, word-level evaluation considers a
word in a predicted span as correct if it is also a word in a gold span. Put
simply, this evaluation gives a positive score to two overlapping spans whose
boundaries do not match exactly.

We use precision, recall, and F\textsubscript{1} scores as evaluation metrics.
We use the evaluation script\footnote{Downloadable from
  \url{https://www.cnts.ua.ac.be/conll2000/chunking/output.html}.} of CoNLL-2000
chunking shared task~\citep{tjongkimsang2000} to perform both types of
evaluation.\footnote{Word-level evaluation is achieved by passing \texttt{-r} as
  option.}

\subsubsection{Majority-Voted and Best-Matched Gold Spans}

We perform strict
majority voting to get the majority-voted gold spans for evaluation. For example, if there are 2 annotators with the following span annotations:
\begin{enumerate}
\item \textit{[I was fired from work] because of [my complaint against my boss] months ago},
\item \textit{I was [fired from work] because of my [complaint against my boss months ago]}
\end{enumerate}
where square brackets denote a span, then the gold
spans are \textit{fired from work} and \textit{complaint against my boss}. In
other words, only words voted by more than 50\% of the annotators are included.
In particular, words voted by exactly 50\% of the annotators are \emph{not}
included.

Another type of gold spans we experiment with is the best-matched spans. Given an input and its predicted spans, best-matched spans of that input are its span annotations against which the predicted spans result in the highest F\textsubscript{1} score when evaluated. These span annotations must come from a single annotator. For instance, if (a)~there are 2 annotators with the same span annotations as before, (b)~the predicted span is only \textit{fired from work}, and (c)~span-level F\textsubscript{1} is used, then the best-matched spans are the spans given by the second annotator. A similar approach has been used in automatic text summarisation~\citep{lin2004}.

\subsection{Model}

We parameterise the function $f$ in \Cref{eqn:prob-spans} with a
neural sequence tagger. The tagger uses a pretrained language model to provide
contextual word representations and a bidirectional LSTM~\citep{hochreiter1997}
with a CRF output layer~\citep{collobert2011} as the classifier similar to
previous work~\citep{lample2016}. We use the implementation provided by the
open-source NLP library FLAIR~\citep{akbik2019}.\footnote{Version 0.13.}

Following prior
work on a similar corpus~\citep{mistica2021}, we use the base and uncased
version of BERT~\citep{devlin2019} as the pretrained language model. The problem
description and the area of law are joined and given as a single text input to
BERT.\@ For example, if the problem description is \textit{My landlord kicked me
  out without reason} and the area of law is \textsc{Housing and residential
  tenancies} then the input is \textit{My landlord kicked me out without reason}
\texttt{<sep>} \textsc{Housing and residential tenancies} where \texttt{<sep>}
marks the end of the problem description. Both \texttt{<sep>} and succeeding input words corresponding to the area of law are excluded from evaluation.

\subsection{Training}

We experiment with two approaches to dealing with subjectivity in model training. The first approach~(\mv{}) aggregates span annotations with majority voting similar to how the majority-voted gold spans are constructed~(\Cref{sec:subj-eval}). This approach resolves subjectivity by only including spans on which the majority of annotators agree.

The second approach is repeated labelling~(\rel{}) which treats multiple
annotations of the same input as separate labelled examples~\citep{sheng2008}.
In other words, annotations in the training set are left as they are without any
attempt to aggregate them. This approach embraces subjectivity by treating all
annotations equally.

While \rel{} may seem counterintuitive because the same input can be
presented with different annotations, these annotations may have consistent
patterns. Spans that are often~(resp.\ rarely) annotated give a strong signal of
the presence~(resp.\ absence) of a true span. We expect that models can learn
the correct spans from these signals.

For both approaches, the tagger is trained for 10 epochs to maximise the probability of the
sequence of tags in the training set. Both learning rate and
batch size are tuned on the development set. The word-level F\textsubscript{1} score against majority-voted spans is used as the hyperparameter tuning objective.

\subsection{Comparisons}

\paragraph{Baseline}

We employ a model that predicts spans randomly as a baseline~(\random{}) which reflects a model that does not perform any learning from data. The model tags each word in the input description with one out of 3 possibilities uniformly at random: start of a span, continuation of a span, or outside of any span. This sequence of tags is then decoded into a set of spans as the output.

\paragraph{Expert performance}

The majority-voted gold spans in \Cref{sec:subj-eval} may not resemble spans
produced by a real annotator. Therefore, even an expert annotator may not achieve
perfect performance when evaluated against the majority-voted gold spans. We
compute this expert performance to serve as a more realistic upper bound of model
performance on our dataset. We estimate this performance by evaluating the
performance of the best annotator of each test input, where best is defined
as resulting in the highest F\textsubscript{1} score against the majority-voted
gold spans. Note that this is different from the best-matched spans mentioned in
\Cref{sec:subj-eval} because here the gold spans are fixed to the majority-voted
spans while the predicted spans come from the best annotator. While there are
limitations to this estimation~(see Limitations), we argue that the
estimate is still useful as a point of reference.

\section{Results}

\begin{table*}\small
    \sisetup{detect-all=true, table-format=2.1(1), separate-uncertainty=true, retain-zero-uncertainty=true}
    \centering
    \begin{subtable}{0.9\textwidth}
        \centering
        \begin{tabular}{@{}l*{6}{S}@{}}
        \toprule
        \multicolumn{1}{c}{\multirow{2.6}{*}{Method}} & \multicolumn{3}{c}{Span} & \multicolumn{3}{c}{Word} \\
        \cmidrule(lr){2-4} \cmidrule(l){5-7}
                  & {P}                & {R}               & {F\textsubscript{1}} & {P}               & {R}               & {F\textsubscript{1}} \\
        \midrule
        \random{} & 0.2(0)             & 4.1(1)            & 0.4(0)               & 17.1(0)           & \bfseries 66.6(0) & 27.2(0)              \\
        \mv{}     & \bfseries 17.9(19) & \bfseries 18.5(3) & \bfseries 18.2(11)   & \bfseries 58.2(4) & 48.7(1)           & \bfseries 53.0(1)    \\
        \rel{}    & 11.2(14)           & 12.6(3)           & 11.8(9)              & 57.5(7)           & 48.9(12)          & 52.8(5)              \\
        \midrule
        Expert    & 80.2               & 67.5              & 73.3                 & 91.0              & 97.5              & 94.2                 \\
        \bottomrule
        \end{tabular}
        \caption{Majority-voted gold spans}\label{tbl:results-mv}
    \end{subtable}

    \vspace{1em}
    \begin{subtable}{0.9\textwidth}
        \centering
        \begin{tabular}{@{}l*{6}{S}@{}}
        \toprule
        \multicolumn{1}{c}{\multirow{2.6}{*}{Method}} & \multicolumn{3}{c}{Span} & \multicolumn{3}{c}{Word} \\
        \cmidrule(lr){2-4} \cmidrule(l){5-7}
                  & {P}                & {R}               & {F\textsubscript{1}} & {P}               & {R}               & {F\textsubscript{1}} \\
        \midrule
        \random{} & 0.0(0)             & 1.2(0)            & 0.1(0)               & 31.0(0)           & \bfseries 66.9(0) & 42.4(0)              \\
        \mv{}     & \bfseries 20.9(22) & \bfseries 26.3(4) & \bfseries 23.3(15)   & 69.2(5)           & 48.6(2)           & 57.1(1)              \\
        \rel{}    & 17.1(22)           & 24.1(5)           & 19.9(17)             & \bfseries 69.6(5) & 48.7(12)          & \bfseries 57.3(7)    \\
        \bottomrule
        \end{tabular}
        \caption{Best-matched gold spans}\label{tbl:results-bm}
    \end{subtable}
    \caption{Span- and word-level precision, recall, and F\textsubscript{1} scores~(in \%) of the span prediction model against majority-voted and best-matched gold spans. Mean~($\pm$ std) across 3 runs are reported except for Expert.}\label{tbl:results}
\end{table*}

\Cref{tbl:results} shows that both \mv{} and \rel{} perform substantially better than \random{} in terms of F\textsubscript{1} scores for all 4 evaluation setups, indicating the potential of both methods. Comparing \mv{} and \rel{} across both types of gold spans, while the former is on par with the latter in word-level evaluation, \mv{} outperforms \rel{} substantially in span-level evaluation. This finding is consistent across precision and recall, and thus demonstrates that \mv{} is overall superior to \rel{}. However, the table also shows that \random{} outperforms both \mv{} and \rel{} in terms of word-level recall across both types of gold spans, which points to an area for improvement.

While the performance numbers with majority-voted gold spans are lower than the best-matched counterparts, the patterns of model performance are consistent across both types of gold spans. This result suggest that both types of gold spans are equally acceptable for handling subjectivity in span annotations. However, using the majority-voted gold spans has the advantage of time efficiency because the gold spans do not need to be recomputed when evaluating different models.

For majority-voted spans, \Cref{tbl:results-mv} shows that model performance is still far behind expert performance, suggesting that there is still plenty of room for improvement. Furthermore, the expert performance is moderately high in span-level evaluation and approaches perfect performance in the word-level counterpart. This finding demonstrates that the majority-voted spans are realistic as they show a high degree of similarity to span annotations given by experts.

\subsection{Experiments with Other Pretrained Language Models}

We also experiment with an improved version of BERT known as
DeBERTaV3~\citep{he2023}. Key differences include a more complex model
architecture, a simpler pretraining objective, and a larger amount of
pretraining data. We use the base version of DeBERTaV3 which has the same number
of layers, attention heads, and hidden units but four times the vocabulary
size of the base version of BERT, as used in the previous experiment. We evaluate
only against the majority-voted gold spans based on the previous findings. Due
to time constraints, we use the hyperparameters~(learning rate and batch size)
tuned on the first fold~(out of 20) for all the folds of the dataset.

\Cref{tbl:deberta-results} shows that both \mv{} and \rel{} outperform \random{}
substantially on both span- and word-level evaluations across all metrics except
for word-level recall where \random{} achieves the best score. This finding
agrees with that of the BERT-based models. Looking at F\textsubscript{1} scores,
the table shows that \rel{} is on par with \mv{} in span-level evaluation and
marginally outperforms \mv{} in the word-level counterpart. This finding
contradicts the results for BERT-based models, suggesting the effectiveness of
\rel{} with improved language models.

Furthermore, the table shows that for span-level evaluation, \rel{} outperforms
\mv{} in precision but performs worse than \mv{} in recall. In contrast, for
word-level evaluation, \mv{} outperforms \rel{} in precision but performs worse
than \rel{} in recall. These findings suggest that with stronger language
models, the best method depends not only on whether span- or word-level
evaluation is prioritised but also on whether precision or recall is more
crucial. These patterns of performance again contradict those of the BERT-based
models, suggesting that the choice of pretrained language models is important.
We leave the analysis on the possible reasons behind these findings and the
evaluation on best-matched gold spans for future work.

Lastly, comparing to \Cref{tbl:results-mv}, we see that DeBERTa-based models
outperform the BERT-based counterparts across the board. This finding is
unsurprising because DeBERTa was developed as an improvement over
BERT~\citep{he2021a}.

\begin{table*}\small
  \sisetup{detect-all=true, table-format=2.1(1), separate-uncertainty=true, retain-zero-uncertainty=true}
  \centering
  \begin{tabular}{@{}l*{6}{S}@{}}
    \toprule
    \multicolumn{1}{c}{\multirow{2.6}{*}{Method}} & \multicolumn{3}{c}{Span} & \multicolumn{3}{c}{Word} \\
    \cmidrule(lr){2-4} \cmidrule(l){5-7}
              & {P}                & {R}               & {F\textsubscript{1}} & {P}               & {R}               & {F\textsubscript{1}} \\
    \midrule
    \random{} & 0.2(0)             & 4.1(1)            & 0.4(0)               & 17.1(0)           & \bfseries 66.6(0) & 27.2(0)              \\
    \mv{}     & 18.4(16)           & \bfseries 19.7(3) & \bfseries 19.0(10)   & \bfseries 61.3(4) & 50.2(3)           & 55.2(0)              \\
    \rel{}    & \bfseries 23.7(26) & 14.8(1)           & 18.2(8)              & 58.7(3)           & 53.0(4)           & \bfseries 55.7(2)    \\
    \bottomrule
  \end{tabular}
  \caption{Span- and word-level precision, recall, and F\textsubscript{1}
    scores~(in \%) of the DeBERTaV3-based model against majority-voted gold
    spans. Mean~($\pm$ std) across 3 runs are reported.\ \random{} performance
    is copied from \Cref{tbl:results-mv}.}\label{tbl:deberta-results}
\end{table*}

\section{Related Work}

\citet{pruthi2020} have studied the span prediction problem under the name of
evidence extraction. However, their model also performs classification jointly
and is trained in a semi-supervised manner. More importantly, they did not
consider subjectivity in the span annotations. In contrast, we focus only on
predicting spans, supervised learning, and incorporating subjectivity in
model training and evaluation.

Previous work has leveraged a similar dataset of legal problem
descriptions~\citep{mistica2021}. They focussed on the text classification
aspect where areas of law are assigned to a problem description. Different from
their work, ours treats the area of law as given and focusses on predicting the
spans that support the assignment of the area of law.

Our work falls within the broader theme of human label
variation~\citep{plank2022}. Previous work has mainly focussed on text
classification tasks~\citep[][\textit{inter
  alia}]{leonardelli2023,fornaciari2021,nie2020}. In contrast, we focus on
spans, which are still understudied in this area. Our work is also related to
data perspectivism.\footnote{\url{https://pdai.info/}}

\section{Conclusion}

We explore the task of automatically predicting text spans in a legal problem
description that support the labelling of an area of law. We develop neural
sequence taggers that deal with the inherent subjectivity of the task.
Experiments across various subjectivity-aware evaluation setups show that
training on majority-voted annotations outperforms training on the disaggregated
counterparts.

\section*{Limitations}\label{sec:limits}

The dataset we use in this work cannot be released publicly, which is a major
limitation of our work in terms of reproducibility. This is because the topics discussed
are sensitive, and more importantly, the help-seekers have not given their
consent to share their data. Nevertheless, we believe our work still offers
valuable scientific knowledge on handling subjectivity, especially in span
annotation tasks.

For the evaluation using majority-voted gold spans, we estimate the expert
performance by determining the best annotator of each test input. However,
the majority-voted gold spans are a function of the best annotator's spans.
Thus, the estimated expert performance is dominated by test inputs that are annotated
by fewer annotators. To mitigate this issue, a leave-one-annotator-out strategy
can be employed, which we leave for future work.

The best-matched gold spans are likely to come from various annotators. Taken
together, these spans may not reflect a realistic pattern of a single human
annotator. A remedy is to evaluate against a single best annotator. However,
this approach is not straightforward in our case because an annotator may
annotate only a subset of examples. We thus leave this approach for future work.

\section*{Acknowledgements}

We thank the anonymous reviewers for their constructive feedback on the paper.
This research is supported by the Australian Research Council Linkage Project
(project number: LP210200917) and funded by the Australian Government. This
research is done in collaboration with Justice Connect, an Australian public
benevolent institution.\footnote{As defined by the \anon{Australian
    government:~\url{https://www.acnc.gov.au/charity/charities/4a24f21a-38af-e811-a95e-000d3ad24c60/profile}}}

\bibliography{span}

\end{document}